# Handwriting Profiling using Generative Adversarial Networks

**Arna Ghosh**[*] and **Biswarup Bhattacharya**[*] and **Somnath Basu Roy Chowdhury**[*]
Department of Electrical Engineering, Indian Institute of Technology Kharagpur
{arnaghosh, biswarup}@iitkgp.ac.in, brcsomnath@ee.iitkgp.ernet.in

## Abstract

Handwriting is a skill learned by humans from a very early age. The ability to develop one's own unique handwriting as well as mimic another person's handwriting is a task learned by the brain with practice. This paper deals with this very problem where an intelligent system tries to learn the handwriting of an entity using Generative Adversarial Networks (GANs). We propose a modified architecture of DCGAN (Radford, Metz, and Chintala 2015) to achieve this. We also discuss about applying reinforcement learning techniques to achieve faster learning. Our algorithm hopes to give new insights in this area and its uses include identification of forged documents, signature verification, computer generated art, digitization of documents among others. Our early implementation of the algorithm illustrates a good performance with MNIST datasets.

## Introduction

Developing a style of handwriting is a relatively complex task for human beings. It is learned from a very young age and is mastered with practice. In fact, the skill to mimic one's own handwriting is not infallible. A person's style of writing has slight variations depending upon the conditions. Therefore, we believe that using Generative Adversarial Networks to artificially synthesize a style of handwriting could offer a better solution to achieve human-like writing. Present state of handwriting recognition is done using recurrent neural networks (Graves et al. 2008). We present a more biologically inspired outlook.

We aim to provide the network a training environment similar to that of humans. We provide the dataset consisting of alphabets (A-Z and a-z) to the generator network to be able to learn to generate individual alphabets. With a reasonably well-trained generator, we use the generator to generate basic words. The initial words generated would be spaced letters, which is what we desire as it is similar to the human behavior. We aim to improve upon the generation of words using the technique of reinforcement learning. The generator learns to generate words looking similar to the reference word provided. Also, we realize that the discriminator network can be used as an OCR (optical character recognition) system for the specific handwriting technique. This acts as an added benefit as we can use the discriminator to provide feedback to the generator while developing words from individual alphabets regarding the desired style. Once again, we believe it is closely linked to the way the human brain works in daily life. With every word we write, we tend to adjust our writing style to correct any difference from the reference - our desired style of writing. Therefore, we expect the algorithm to provide human-like results. Extending this idea, similar networks can be trained to learn other complicated tasks which require practice, like playing a sport, generating music etc. The implementation details and the network specifications along with experiments performed are laid out in the further sections.

## Architecture

We have used the popular architecture of a stable deep convolutional generative adversarial network (DCGAN) (Radford, Metz, and Chintala 2015). The initial layers of the DCGAN consist of a convolutional network which performs strided convolution, replacing the spatial sampling in conventional CNNs. This aids the generator depicted in Figure 1 to learn its own upsampling. The same convolutional layers are also present in the discriminator in reverse order. The fully connected layers are absent in DCGAN. The features extracted from the highest convolutional layer form the input of the generator and output of the discriminator. In order to stabilize the learning process, batch normalization is applied to the discriminator input. Batch normalization is used for normalizing the input to have a zero mean and unit variance.

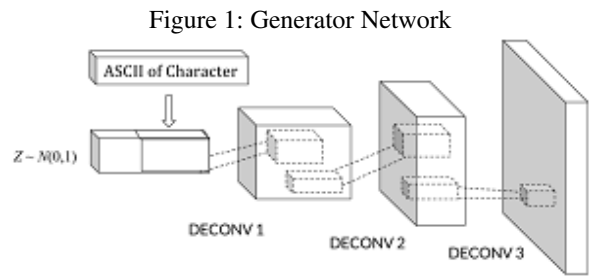

Figure 1: Generator Network

The generator and discriminator architecture are inspired from (Reed et al. 2016). The ASCII value of the character to be generated is concatenated with a noise vector and provided as input to the generator as shown in Figure 2. Feedforward through a normal deconvolutional network in generator $G$ results in a synthetic image $\tilde{x}$. The discriminator $D$ has a convolution network followed by leaky ReLU. We reduce the dimensionality of the image and finally concatenate the character embedding to the output to calculate the score from $D$.

---

[*]Equal Contribution

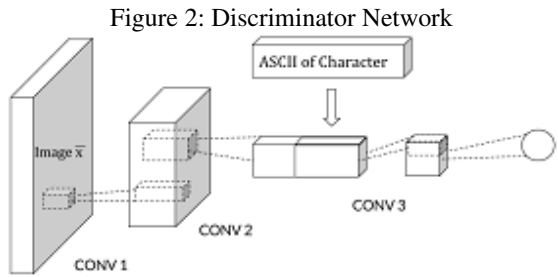

Figure 2: Discriminator Network

The easiest way to train the GAN is to view {character, image} pairs as joint observations and train the discriminator to judge pairs as real or fake. However, this type of conditioning does not allow the discriminator to learn whether real training images match the character embedding. Therefore, similar to (Reed et al. 2016), in addition to the real or fake inputs to the discriminator during training, we add a third type of input consisting of real images with incorrect character ASCII value, which the discriminator must learn to score as fake.

## Integrating Reinforcement Learning

The notion of reinforcement learning comes into play when letters need to be joined to form words. For example, the spacing between characters, strokes from one letter to another in cursive handwriting are generally unique to a handwriting style. Hence, the idea is to set up a reinforcement learning technique to give appropriate rewards or penalties so that the generator is able to learn the handwriting with greater accuracy.

The idea of reinforcement learning is applied to the average distance between letters, angle of ascent and angle of descent of strokes. On training, the average distances (space) between the letters in words are tracked. Like this, 26x26x2 values are obtained (letter combinations can be Aa, Ab, ..., Ba, Bb, ..., Zy, Zz, aa, ab, ..., ba, bb, ..., zy, zz). Letter combinations like aA, FF etc. are ignored as such continuous combinations do not occur in real English words. Similar rules can be easily created for any given language. The angle of ascent and descent of strokes between characters are also similarly noted. The angles are calculated using OCR techniques (Arica and Yarman-Vural 2002).

Our system exclusively uses a penalty-based system. The difference between the spacing between two letters and the average spacing between those two letters is taken to be an objective function which needs to be minimized. A similar objective function exists for the angles (one for ascent and one for descent). The GAN system has a controller which keeps updating (reinforcing) the thresholds according to the result of this objective function. The policy of the system is to achieve zero as the result of all the objective functions indicating perfectly following expected profile. The penalty system is expected to converge to this policy given enough training data.

## Experimental Analysis

We ran our preliminary experiments with the MNIST dataset. Instead of using the character dataset of one's handwriting, we use the handwritten digit database to test the scope of our architecture in generating digits. We correspondingly use the ASCII value of digits 0 to 9 for the purpose. The wide variation of the style in writing leads to variations in generated images of the same digit. For a particular style of writing, we believe that such variations will be low and the generator network will be far more accurate in replicating a particular style of writing.

Further experiments would include running on a particular handwriting dataset. We plan to use the sentence "the quick brown fox jumped over the lazy dog" written multiple times to generate a dataset covering most of the variations. Further work would involve using trained generator and discriminator to generate words. The dataset for the same will be generated accordingly.

## Conclusion

Our algorithm hopes to provide some insight into the role of GANs in attaining human-like behavior in complex activities and the combination of GANs with existing techniques like reinforcement learning to improve performance. Its uses include, but are not limited to, identification of forged documents, signature verification, computer generated art, digitization of documents. The implications of the results of the handwriting task are far-reaching. The task of mimicking a handwriting style is just an illustration that a complex task that requires practice to be mastered can be learned using the proposed architecture and methodology. Therefore, the main purpose of the paper is to present a technique that we believe is novel to the best of our knowledge and closely mimics the learning network in human beings.